\documentclass[twocolumn]{svjour3}         
\smartqed  
\usepackage{times}
\usepackage{epsfig}
\usepackage{graphicx}
\usepackage{amsmath}
\usepackage{amssymb}
\usepackage{color}
\usepackage{threeparttable}
\usepackage{multirow}
\usepackage{booktabs}
\usepackage{comment}
\usepackage{pbox}
\usepackage[]{natbib}
\usepackage[pagebackref=true,breaklinks=true,letterpaper=true,colorlinks,bookmarks=false]{hyperref}

\newcommand{\etal}{\textit{et al}.}
\newcommand{\ie}{\textit{i}.\textit{e}.~}
\newcommand{\eg}{\textit{e}.\textit{g}.}
\newcommand{\etc}{\textit{etc}.}

\newcommand{\ieno}{\textit{i}.\textit{e}.}

\begin{document}

\title{Target-Tailored Source-Transformation for Scene Graph Generation
}

\titlerunning{TTST}        

\author{Wentong Liao{$^{1}$}       \and
        Cuiling Lan{$^2$} \and
        Wenjun Zeng{$^2$} \and
        Michael Ying Yang{$^3$} \and
        Bodo Rosenhahn{$^1$}
}


 \authorrunning{Liao \etal} 

\institute{Wentong Liao \at
              liao@tnt.uni-hannover.de          
           \and
           Cuiling Lan \at
              culan@microsoft.com
           \and
           Wenjun Zeng \at
              wezeng@microsoft.com
           \and
           Michael Ying Yang \at
              michael.yang@utwente.nl
           \and
           Bodo Rosenhahn \at
              rosenhahn@tnt.uni-hannover.de
            \and
        	$^1$ \quad Institute f\"ur Informationsverarbeitung, Leibniz Universit\"at Hannover, Germany \at
        	$^2$ \quad Microsoft Research Asia, Beijing, China \at
        	$^3$ \quad Scene Understanding Group, University of Twente, The Netherlands \at
}

\date{Received: date / Accepted: date}

\maketitle
\begin{abstract}
Scene graph generation aims to provide a semantic and structural description of an image, denoting the objects (with nodes) and their relationships (with edges).
The best performing works to date are based on exploiting the context surrounding objects or relations, \eg, by passing information among objects.
In these approaches, to transform the representation of source objects is a critical process for extracting information for the use by target objects.
In this work, we argue that a source object should give what target object needs and give different objects different information rather than contributing common information to all targets. To achieve this goal, we propose a Target-Tailored Source-Transformation (TTST) method to efficiently propagate information among object proposals and relations.
Particularly, for a source object proposal which will contribute information to other target objects, we transform the source object feature to the target object feature domain by simultaneously taking both the source and target into account.
We further explore more powerful representation by integrating language prior with visual context in the transformation for scene graph generation.
By doing so the target object is able to extract target-specific information from the source object and source relation accordingly to refine its representation.
Our framework is validated on the Visual Genome benchmark and demonstrated its state-of-the-art performance for the scene graph generation.
The experimental results show that the performance of object detection and visual relationship detection are promoted mutually by our method.

\keywords{Scene graph generation \and Message passing \and Feature transformation}
\end{abstract}

\begin{figure*}
\begin{center}
\includegraphics[width=0.9\linewidth]{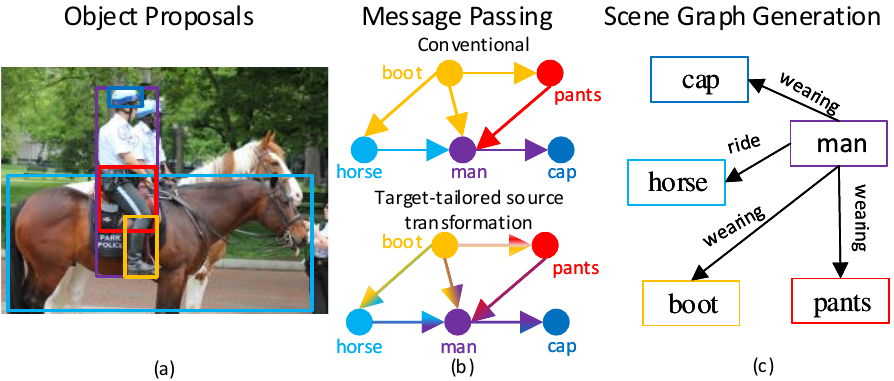}
\caption{Given an image: (a) objects are proposed; (b) Messages are passed among objects to exploit context; (c) label the graph's nodes (objects) and edges (relations). In (b), we show the difference between the conventional message passing methods (top graph) in which a source object always contributes identical representation to different target objects while our target-tailored source-transformation (bottom graph) enables a source object contributes different information to different target objects, by jointly considering the source object and the target object. The arrows denote the message passing direction, the colors indicate the passed information. Different colors indicate the corresponding objects.
}
\label{fig:defineation}
\end{center}
\end{figure*}
\section{Introduction}
\label{sec:intro}
In recent years great successes have been witnessed on vision perceptual tasks such as image classification \citep{perronninomas2010improving,simonyan2014very,szegedy2015going}, object detection \citep{felzenszwalb2009object,girshick2015fast,ren2015faster,liu2016ssd,he2017mask,redmon2017yolo9000}, semantic segmentation \citep{silberman2012indoor,long2015fully,zhao2017pyramid}.
However, these object-centric visual perception is still far from the goal of visual scene understanding which requires understanding the visual relationships between objects \citep{papalia2007human,firestone2016cognition}.

Some recent work \citep{johnson2015image,krishna2017visual} proposed to represent the visual scene as a scene graph which models objects and their attributes as nodes, and their relationships as edges, as illustrated in Fig.~\ref{fig:defineation}.
Scene graph has been proved to be a promising alternative for many visual tasks such as image retrieval~\citep{johnson2015image}, image caption \citep{yao2018exploring,gao2018image,xu2019scene,yao2019hierarchy,yang2019auto}, visual question and answering \citep{johnson2017clevr,teney2017graph,wu2017image,li2019relation}, image generation \citep{johnson2018image,yikang2019pastegan,zhao2019image,ashual2019specifying}, etc. The task of scene graph generation has been attracting increasing attention \citep{silberman2012indoor,lu2016visual,yang2017support,qi2018learning,gkioxari2018detecting,zellers2018neural,liao2019natural,hu2019exploiting}.

A natural idea to generate scene graph is to detect objects using an off-the-shelf object detector, and then predict their pairwise relationships \emph{separately}~\citep{lu2016visual,mallya2016learning,zhang2017visual}.
However, these approaches ignore the exploration of visual context, which could provide powerful inductive bias and strong regularities \citep{zellers2018neural} that help detect objects and reason their relations.
For example, ``keyboard" and ``mouse" often co-occur within a scene, and  ``man" tends to ``ride" the ``horse".
Many works have exploited the visual context in different ways to help scene graph generation \citep{xu2017scene,yu2017visual,li2017vip,li2017scene,dai2017detecting,zellers2018neural}.
Particularly, modeling message passing among objects is the most widely applied method for exploiting the visual context and its effectiveness has been proved for scene graph generation.
In previous message passing methods, the representation of a source object is first transformed, via a learned \emph{shared transformation} $W$ before being remedied to update the target object \citep{xu2015show,li2018factorizable,li2017scene,yang2018graph}.
To make the shared transformation suitable for any target object, $W$ is unfortunately encouraged to learn information from source objects which is commonly useful for different target objects.

\begin{figure*}[ht]
\centering
\includegraphics[width=1\linewidth]{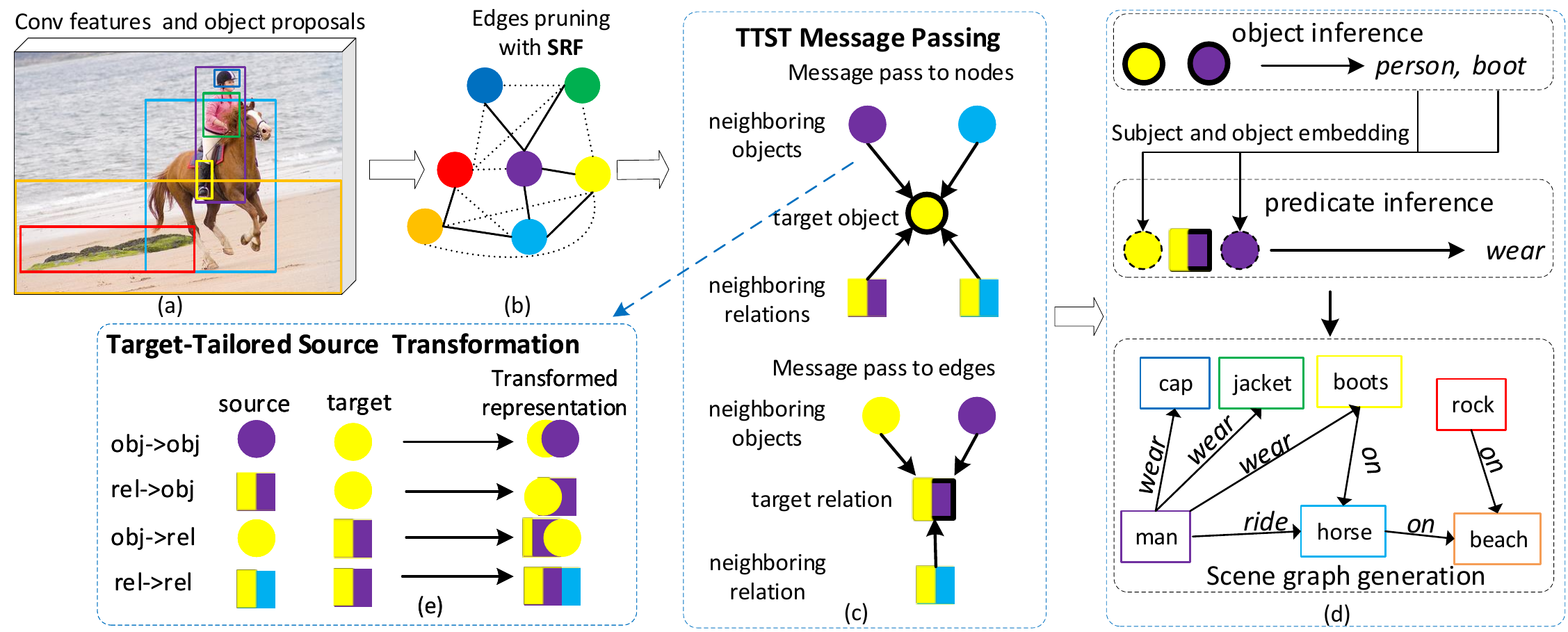}
 \caption{The pipeline of our framework. Given an image,
(a) Faster R-CNN is implemented to propose object candidates and extract visual features,
(b) then our semantic relation filter (SRF) prunes the connection between a pair of objects that are semantically weakly dependent (pointed lines).
(c) Target-tailored source-transformation (depicted in (e)) is applied to learn context from connected nodes and edges across the graph. 
(4) After predicting the objects using the refined object features, their predicted label information are embedded to serve as subject or object in a relationship for relationships inferring.
Finally, scene graph is generated.
The colors indicate different objects. Circles denote objects and rectangles denote relationships.
}
\label{fig:framework}
\end{figure*}

We argue that two important elements are overlooked in most of the existing message passing methods for scene graph generation.
First, the semantic dependencies between target objects and source objects are ignored in the source-transformation step because this transformation is independent of the target object.

For example as shown in Fig. \ref{fig:defineation}(b)(top), a ``boot" will contribute the same information to ``person" as to ''horse" (edges are denoted by the same color) with the shared transformation.
Intuitively, target-tailored information is more useful than common information for a specific target object.
For example, ``boot" should contribute information of ``wearing things" to ``person" while contributing information of ``riding gear" to ``horse" (See Fig. \ref{fig:defineation}(b)(bottom) edges are denoted by different colors).
Second, how to effectively couple the visual and language context into the learning process has not attracted much attention.
The visual appearance determines the visual context while the language prior guides how objects relate to each other in the linguistic domain.
For instance, when we see ``person on a motor" (visual context), we humans spontaneously infer the relation as ``ride"  rather than ``on" or ``sit" (language prior).
The language prior and visual information should be compatible and mutually promotive rather than implemented separately.

Motivated by these observations, we propose a \emph{target-tailored source-transformation} (TTST) method for message passing to exploit context for scene graph generation.
``Target-tailored" means that when a source contributes information to different targets, we expect it to deliver target specific information, \ieno, give as exactly as possible what the target needs. To achieve this goal, we simultaneously consider the source and target when transforming the source information to the target domain. 
Furthermore, we propose to integrate the language priors with visual context in the transformation process. 
By doing so, messages are propagated through the graph more effectively and the learned representations are more powerful.

We depict our framework in Fig.~\ref{fig:framework}. We build it based on the Faster R-CNN detector~\citep{ren2015faster} to generate object proposals (Fig.~\ref{fig:framework}(a)). Then, a graph is initialized by connecting each pair of objects. We introduce a learned semantic relationship filter (\textbf{SRF}, see Sec.~\ref{subsec:srf}) to prune the spurious connections between objects (Fig.~\ref{fig:framework}(b)) to facilitate the subsequent message passing processes.
Then, we apply the proposed target-tailored source transformation for message passing from connected nodes and edges in the graph (see Sec.~\ref{subsec:stamp}, Fig.~\ref{fig:framework}(c)(e)). 
Finally, the labels of graph nodes are predicted with the context-rich features, and the edge labels are inferred by using the refined relationship features along with the semantic information of the connected object nodes (Fig.~\ref{fig:framework}(d)).



Our work has two major contributions:
\begin{itemize}
    \item We propose an effective target-tailored source-transfor-\\mation method for message passing, which explores information from source object/relation to refine target object/relation by considering the source object/relation and target object/relation simultaneously.
    \item Language context is utilized to help message passing for learning powerful representation for scene graph generation.
\end{itemize}

Our framework achieves the state-of-the-art results on the VG benchmark dataset \cite{krishna2017visual} for scene graph generation. Moreover, the experimental results demonstrate the mutual improvements of object detection and relationship detection via our method.


\section{Related Works}
\label{sec:related_work}


\textbf{Context for Visual Reasoning.} Context has been explored to improve different scene understanding tasks for decades \citep{divvala2009empirical,ladicky2010graph,yao2010modeling,hu2018relation,liu2018structure}.
\cite{silberman2012indoor} proposed to infer the support relation between segmented objects, and utilize the interaction context to improve the performance of object segmentation in the indoor scenes. \cite{yang2017support} proposed to generate a scene graph for each image by reasoning the support relations between objects and using the scene context.
To learn better contextual information, a number of works attempt to capture object context from an image in the message passing mechanism, such as through a graph model \citep{li2018factorizable,yang2018graph}, implementing RNN \citep{zellers2018neural,chen2018scene,wang2019exploring,hu2019exploiting}, or in an iterative refinement process \citep{xu2017scene}.

Besides visual context, contexual information from language priors~\citep{mikolov2013efficient,pennington2014glove} has been proved to be helpful for visual relationships detection and scene graph generation \citep{lu2016visual,yu2017visual,li2017scene,liao2019natural}.
Lu \etal \citep{lu2016visual} made use of language priors to improve the detection of meaningful relationships between objects.
Li \etal \citep{li2017scene} exploited language priors from region captions for scene graph generation by predicting image caption and detecting visual relationships in parallel.
Yu \etal \citep{yu2017visual} distilled linguistic knowledge by training a parallel language branch as a teacher network to help the visual network (student) predict visual relationships.
Liao \etal \citep{liao2019natural} proposed to use the language prior from pre-trained word2vector to guide the model to infer the relationship between objects belonging specific categories.

In contrast to above works that utilize language prior separately, we integrate the language priors and the visual context in the transformation step to help message passing and learn better semantic representations.

\textbf{Scene Graph Generation.}
Scene graph was first proposed in \citep{johnson2015image} and implemented for image retrieval. 
It generalizes the task of detecting object to also detecting their attributes and reasoning relationships between them.
Scene graph generation which includes object detection and visual relationship detection are attracting increasing attention in computer vision \citep{li2017scene,dai2017detecting,liang2017deep,zhuang2017towards,li2018factorizable,zellers2018neural,woo2018linknet,wang2019exploring,chen2018scene,chen2019knowledge}.
Context has been proved to be useful for scene graph generation and many works resort to message passing to exploit the contextual information of the related objects \citep{li2017vip,zellers2018neural,chen2019knowledge,wang2019exploring,liao2019natural}, or of the objects and their relationships \citep{xu2017scene,li2017scene,li2018factorizable,yang2018graph,hu2019exploiting}.

Nevertheless, all existing transformation methods do not take the target into account.
Consequently, to any target, the source contributes identical information. For instance, ``horse" contributes the same content to ``human'' and ``grass" after the transformation, even though an attention mechanism is used to weight the contribution.
However, intuitively, the transformed content should be dependent on both the target and source.
Our TTST for message passing is essentially different from previous works by considering the source objects and target object simultaneously. By doing so, for a different target object, the source object contributes different information, and thus the learned representation of target object is more powerful.


\section{Proposed Approach}
\label{sec:approach}
An overview of our proposed model is depicted in Fig. \ref{fig:framework}.
Our goal is to infer a scene graph $G$ for a given image $I$, which summaries the objects $O$ as nodes and relations $R$ between every two objects as edges.
The inferring process can be formally defined as:
\begin{equation}
P(G|I) = P(B|I)P(O|B,I)P(R|B,O_s,O_o,I)
\label{eq:definition}
\end{equation}
where $B$ are locations of objects, $O_s,R,O_o$ stand for subject, relation (predicate), and object, respectively, and $O_s,O_o$ $\in O$. $P$ denotes the inference probability.
$P(B|I)$ can be modeled by an off-the-shelf object detector (Fig. \ref{fig:framework}(a)).
We will discuss each inference module of $P(O|B,I)$ (Fig. \ref{fig:framework}(c)) and $P(R|B,O_s,O_o,I)$ (Fig. \ref{fig:framework} (d)) in the following.

\subsection{Object Proposals}
Given an image, we use Faster R-CNN \citep{ren2015faster} to generate a set of object proposals $O$, as shown Fig. \ref{fig:framework}(a).
Each detected object $o_i\in O$ is associated with its located region $b_i=[x_i,y_i,w_i,h_i] \in B$, initially predicted label distribution over all $C$ classes $p^o_i \in \mathbb{R}^{C}$, and the pooled visual feature vector $x^o_i$.

\subsection{Semantic Relationship Filter}
\label{subsec:srf}
With $n$ object proposals, there are $\mathcal{O}(n^2)$ edges in the fully connected graph when considering every two objects have a relation (Fig. \ref{fig:framework}(b)).
It has been pointed out in many previous works that most of the object pairs have no relationship due to the real-world regularities of objects interaction (dash edges in Fig. \ref{fig:framework}(b)).
We have also observed that information propagated from the unrelated objects could deteriorate the system's performance because of the possible noise and interfering information. On the other hand, message passing through a fully connected graph is computationally costly and of low efficiency.
To make the message passing processes more effective, we propose a \emph{semantic relationship filter} (SRF) to remove the unlikely relationships, similar to what is done in \citep{yang2018graph}.

For an object $o_i$, we compute its semantic representation by multiplying its estimated class distribution by the semantic semantic
word embedding matrix $\mathbf{W}_e$:
\begin{equation}
e^o_i = p^o_i\cdot{\mathbf{W}_e},
\label{eq:embed}
\end{equation}
where each entry in $\mathbf{W}_e$ is an embedding vector for the corresponding object class.
It is learned from the region caption annotation of the Visual Genome (VG) dataset~\citep{krishna2017visual} by adopting Glove \citep{pennington2014glove}.
A multi-layer perception (MLP) is trained to estimate a semantic relatedness score between $o_i$ and $o_j$ by feeding $[e^o_i, \tilde{b}_i, \tilde{b}_j,e^o_j]$, where $[\cdot]$ denotes a concatenation operation and $\tilde{b}_i$ is the normalization of $b_i$ with respect to the union box of $(o_i,o_j)$.
Then, the object pairs with the top $K$ relatedness scores which are also larger than an empirical threshold, are kept and denoted as $R$.

\begin{figure}
    \centering
    \includegraphics[width=1.0\linewidth]{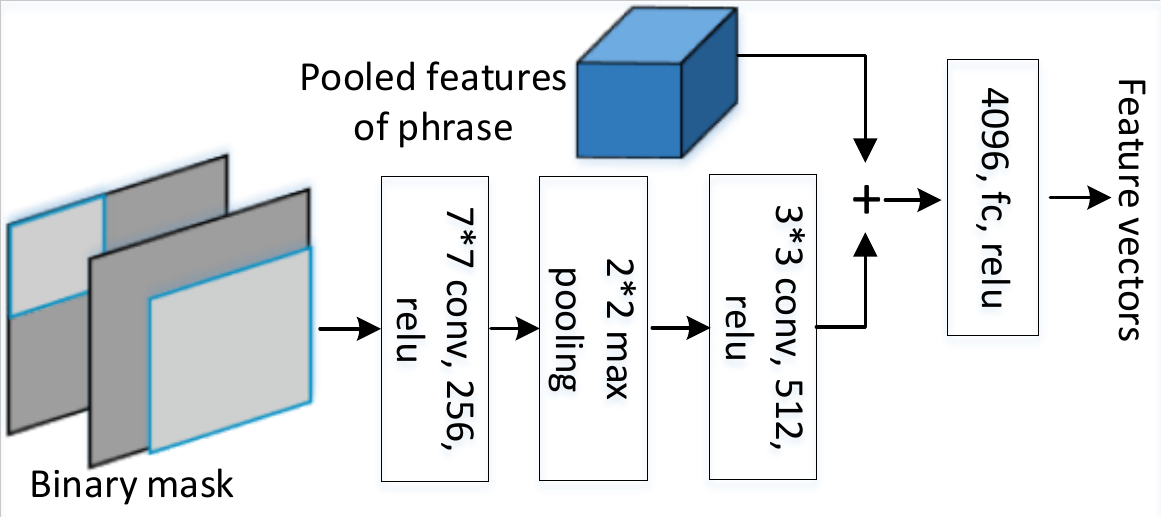}
    \caption{Depiction of the sub-module that extracts relationship features.}
    \label{fig:spatial}
\end{figure}

A relationship of $(o_i,o_j)$ is denoted as $r_{ij}$.
We extract its basic representation $x_{ij}$ by fusing its visual feature with the spatial feature, as depicted in Fig. \ref{fig:spatial}.
The visual feature is pooled from the shared feature maps on the tight union box of $(o_i,o_j)$.
The spatial feature is first represented as a two-channel binary mask, which indicates the places of subject and object respectively. Within the mask the pixels which are within the region of subject/object are denoted as 1, otherwise as 0.
Then they are fed forward to two sequential convolutional layers each of which is followed by a ReLU activation. A max pooling operation is inserted between the two convolutional layers.

\subsection{Target-Tailored Source-Transformation for Message Passing}
\label{subsec:stamp}
\subsubsection{Message Passing Revisited}
Generally, passing message to a target node $i$ from its neighboring nodes $\mathcal{N}(i)$ at the $l+1$ step can be defined as:
\begin{equation}
    z^{l+1}_i =\sigma(z^{l}_i + \Sigma_{j\in\mathcal{N}(i)}a_{ij}Wz^{l}_j),
    \label{eq:mp}
\end{equation}
where $a_{ij}$ is the weight for the neighboring node $j$ and is computed using attention mechanism typically. $W$ is a shared learned transformation matrix which is used to project the representation of source objects to a common domain. $\sigma(\cdot)$ is a nonlinear operation. After several iterations, a representation with a high-order context is obtained and forwarded to the subsequent inference module.

However, $Wz_j$ contributes the same information to any target $z_i$.
Ideally, the transformation should consider the semantic dependency between the target $z_i$ and the source $\mathcal{N}(i)$. 
To address this problem, we propose the \emph{target-tailored source-transformation} (TTST) for message passing to better explore context through the graph. 
The TTST message passing process is depicted in Fig. \ref{fig:framework}(c)(e) and discussed in the following.

\subsubsection{TTST for Objects}
To learn the context of objects and relationships at different semantic levels, messages are passed from both the neighboring objects ${N}^o(i)$ and relationships ${N}^r(i)$ to the target object. This message passing is formulate as:
\begin{equation}
    \begin{aligned}
        \hat{x}_i =\sigma(x_i + & \frac{1}{|\mathcal{N}^o(i)|}\Sigma_{j\in\mathcal{N}^o(i)} f^{(o \rightarrow o)}([x_i,e_i],[x_j,e_j])\\ + &\frac{1}{|\mathcal{N}^r(i)|}\Sigma_{j\in\mathcal{N}^r(i)} f^{(r \rightarrow o)}(x_i,x_{ij})).
    \label{eq:stamp_obj} 
    \end{aligned} 
\end{equation}
Note that, the superscript $l$ is removed for simplicity.
The superscript $o$ and $r$ represent object and relationship respectively.
$f^{(\rightarrow )}(target, source)$ is our TTST operation and the arrow indicates the message passing direction.
It is worth noting that $e_i$ is computed by Eq. \eqref{eq:embed} which contains language prior. It is concatenated with the visual feature $x_i$ as complete representation of object $i$.
Therefore, $f^{(o \rightarrow o)}$ broadcasts the visual information as well as the language prior among object nodes.
Consequently, both the visual context and the language prior between objects are learned and integrated into the refined representations of target objects.
Because the transformation $f^{(\rightarrow )}(\cdot)$ ``sees" the target and object simultaneously, it is target-tailored source-transformation.
Moreover, the transformation is further better guided by the implicit language prior in $e_i$ between different classes of objects.
The ablation studies in Sec.~\ref{sub:ablation} will show how the language prior affects the performance.

\renewcommand{\arraystretch}{1.3} 
\begin{table*}[htp!]
\centering
\caption{Performance comparison with state-of-the-art on VG test set \citep{xu2017scene}. All numbers in \%.
We use the same object detection backbone provided by \citep{zellers2018neural} for fair comparison. Because MSDN, FacNet and DRNet use their own data split, the comparison is for reference only.
The results of VRD are taken from \citep{xu2017scene} which reimplemented VRD on VG dataset. The results of Graph R-CNN, KERN, Mem NLPVR and AVR are taken from the original papers. Because NLPVR uses different experimental settings for task of PredCls, we do not compare them for this task for fairness purpose.}
\label{tab:results}
\begin{threeparttable}
\begin{tabular}{ccccccccccc}
\toprule
\multirow{2}{*}{ } & \multirow{2}{*}{ Method } & \multicolumn{3}{c}{ SGGen } & \multicolumn{3}{c}{ SGCls } & \multicolumn{3}{c}{ PredCls } \cr
    \cmidrule(lr){3-5} \cmidrule(lr){6-8} \cmidrule(lr){9-11}
& &R@20 & R@50 & R@100 & R@20 & R@50 & R@100 & R@20 & R@50 & R@100\cr
\midrule
&VRD \citep{lu2016visual} &  & 0.3 & 0.5 & & 11.8 & 14.1 & & 27.9 & 35.0\\
&IMP \citep{xu2017scene} &14.6  & 20.7 & 24.6 & 31.7 & 34.6 & 35.4 &52.7 &59.3 &61.3\\
&Graph R-CNN \citep{yang2018graph} & - & 11.4 & 13.7 & - & 29.6 & 31.6 & - & 54.2 & 59.1 \\
&Mem \citep{wang2019exploring}  &7.7 & 11.4 &13.9 & 23.3 & 27.8 & 29.5 & 42.1 & 53.2 & 57.9 \\
&NLPVR\citep{liao2019natural}& - &22.0 &23.5 & - & 28.0 & 30.1 &- &- & -\\
&AVR\citep{hu2019exploiting}& - &19.4 &22.7 & - & 29.4 & 34.5 &- &58.2 & 60.7\\
&KERN \citep{chen2019knowledge}  &- & 27.1 &29.8 &- & 36.7 & 37.4 & - & 65.8 & 67.6 \\
&MotifNet-Freq \citep{zellers2018neural} & 20.1 & 26.2 & 30.1 & 29.3 & 32.3 & 34.0 & 53.6 & 60.6 & 62.2 \\
&MotifNet \citep{zellers2018neural} & 21.4 & 27.2 & 30.3 & 32.9 & 35.8 & 36.5 & 58.5 & 65.2 & 67.1 \\
\midrule
\multirow{2}{*}{\rotatebox[origin=c]{90}{Ours}}
& TTST (w/o SRF)  & 22.4 & 29.3 & 33.1 & 34.2 & 37.3 & 38.3 & \textbf{61.8} & \textbf{66.5} & \textbf{67.7} \\
& TTST  & \textbf{23.8} & \textbf{32.3} & \textbf{35.4} & \textbf{35.1} & \textbf{38.6} & \textbf{39.7} & 60.3 & 64.2 & 66.4 \\
\bottomrule
\end{tabular}
\end{threeparttable}
\end{table*}

\subsubsection{TTST for Relationships.}
TTST is also applied to capture context for relationships from its neighboring objects (\ie, the subject and object) and neighboring relationships ${N}^r(i,j)$ as follows: 
\begin{equation}
    \begin{aligned}
        \hat{x}_{ij} =\sigma(&x_{ij} +  \frac{1}{2}\Sigma_{m\in[i,j]} f^{(o \rightarrow r)}(x_{ij},x_m)\\ + &\frac{1}{|\mathcal{N}^r(i,j)|}\Sigma_{x_{nm}\in\mathcal{N}^r(i,j)} f^{(r \rightarrow r)}(x_{ij},x_{nm})).
    \label{eq:stamp_rel} 
    \end{aligned} 
\end{equation}
${N}^r(i,j)$ is defined as the set of relationships in which each relationship involves either $o_i$ or $o_j$.
It is worth noting that, after the first iteration in Eq.~\eqref{eq:stamp_obj}, $x_i$ and $x_j$ contain context of the language prior.
Consequently, $f^{(o \rightarrow r)}(\cdot)$ also integrates context of the language prior to the relationship representation.

Each transformation $f(\cdot)$ is a separately learned MLP (two fully connected (FC) layers followed by a Relu operation). 
Each of them is responsible for passing messages in different directions and capturing different levels of context.

\subsection{Inference}
The inference module is depicted in Fig. \ref{fig:framework}(d).
An object classifier is trained to predict the label distribution $\hat{p}^o_i$ of object proposal $i$ using $\hat{x}_{i}$.
Thus, $P(O|B,I)$ in Eq.~\eqref{eq:definition} is achieved.
To infer the graph edge label (\textit{i.e.} relation class), we semantically embed $\hat{p}^o_i$ and $\hat{p}^o_j$ to further explore context information of (subject, relation, object).
\begin{equation}
    e^{sub} = \hat{p}^o_{sub}\cdot W^{sub}_{emb},~~e^{obj} = \hat{p}^o_{obj}\cdot W^{obj}_{emb}, 
    \label{eq:pred_emb1}
\end{equation}
where $W^{sub}$ and $W^{obj}$ denote the trainable embedding matrix of subject and object respectively. 
Then, the relationship is semantically represented as $\tilde{x}_{ij}=[e^{sub},\hat{x}_{ij},e^{obj}]$.
Different from most of the previous works which simply combine the visual features or predicted label distribution of subject and object with the features of relationship, we further explore their context information.
Finally, an MLP (consisting of two FC layers followed by a Relu and softmax operation sequentially) is trained to predict the relation class distribution using $\tilde{x}_{ij}$.
Now, $P(R|B,O_s,O_o,I)$ in Eq. \eqref{eq:definition} is achieved.
The labels of objects and relations that maximize Eq.~\eqref{eq:definition} are selected.

\section{Experiments}
\label{sec:exp}
In this section we firstly clarify the experimental settings and implementation details.
Then, we compare our methods with the state-of-the-art approaches.
We conduct extensive ablation study on each module of our framework and demonstrate their effectiveness.

\textbf{Datasets.~}
The Visual Genome (VG) dataset~\citep{krishna2017visual} is the largest and most popular benchmark dataset for the task of scene graph generation.
However, different works use different data splits.
For a fair comparison, we adopted the most widely adopted ddataset split in \citep{xu2017scene}.
In the data split, the most-frequent 150 object categories and 50 predicate types are selected.
The dataset is split into a training set with $75651$ images and a test set with $32422$ images.
Visual Relationship Detection (VRD) \citep{lu2016visual} dataset is another popular dataset for relationship detection in the early stage. However, it is a very small subset of the VG dataset, so as the recent works, we will not conduct experiments on it.

\renewcommand{\arraystretch}{1.3} 
\begin{table*}[!th]
\centering
\caption{Ablation studies on our model with accuracy in \%.
\textbf{TTST} denotes whether pass message to capture context through the graph using our proposed TTST message passing method.
\textbf{Language} denotes using the language context in message passing.
\textbf{PredE} denotes the semantic embedding of subject and object of a relationship as defined in Eq.~\eqref{eq:pred_emb1}.
\textbf{SRF} stands for the semantic relationship filter which is trained to prune the spurious edges.
The object detection performance (mAP) follows COCO metrics \citep{lin2014microsoft}.}
\label{tab:ablation}
\begin{threeparttable}
\begin{tabular}{c|cccccccccccc}
\toprule
\multirow{2}{*}{Model} & \multirow{2}{*}{TTST} &\multirow{2}{*}{Language} & \multirow{2}{*}{PredE} & \multirow{2}{*}{SRF}&  Detection &\multicolumn{2}{c}{SGGen} & \multicolumn{2}{c}{SGCls} & \multicolumn{2}{c}{ PredCls } \cr 
 \cmidrule(lr){6-6} \cmidrule(lr){7-8} \cmidrule(lr){9-10} \cmidrule(lr){11-12}
& & & & 								& mAP & R@50 & R@100 & R@50 & R@100 & R@50 & R@100\cr
\midrule
1 &-			& - & - &- 				&16.6 & 12.7 & 15.9 & 26.6 & 27.4 & 52.4 &54.1 \\
2 &\checkmark 	& - &-	& - 			&18.5 & 17.1 & 19.9 & 29.7 & 32.4 & 58.3 &60.4 \\
3 &\checkmark 	& \checkmark & - &- 	&20.2 & 24.7 & 27.1 & 33.0 & 35.1 & 62.0 &64.2 \\
4 &\checkmark &\checkmark &\checkmark &- &20.4 &29.3 &33.1 &37.3 &38.3 &\textbf{66.5} &\textbf{67.7} \\
5 &\checkmark&\checkmark &\checkmark &\checkmark &\textbf{20.8} &\textbf{32.3} &\textbf{35.4} &\textbf{38.6} &\textbf{39.7} &64.2 &66.4 \\
\bottomrule
\end{tabular}
\end{threeparttable}
\end{table*}

\textbf{Implementation Details.~}
Faster R-CNN \citep{ren2015faster} with VGG16 \citep{simonyan2014very} as backbone is implemented as our underlying detector and basic visual feature extractor.
The codebase is provided by \citep{zellers2018neural}.
The input images are scaled and then zero-padded to the size of $592\times 592$.
ROI-pooling \citep{girshick2015fast} is applied to extract features of nodes and edge from the basic shared feature maps.
In the SRF module, the embedding matrix $W_e$ is initialized with the $300$-D (dimensions) Word2vec provided by \citep{lu2016visual}, and a two-layer MLP is trained to output a $1$-D vector which then goes through a sigmoid function to squash the predicted score in $(0,1)$.
We enable SRF to keep at most $128$ relationship proposals with threshold empirically set to $0.55$ by considering the trade-off between high recall and accuracy of correct relationships.
Each feature transformation $f(\cdot)$ in TTST is an MLP which consists of two FC layers (each followed by Relu operation) and outputs $512$-D feature vectors for objects and $4096$-D feature vectors for relationships, respectively.
The embedding matrices $W^{sub}_{emb}, W^{obj}_{emb} \in \mathbb{R}^{50\times 300}$ are randomly initialized, where each row corresponds to an object class.

All experiments are conducted on a single GTX 1080 Ti graphic card with Pytorch framework.

\textbf{Training.~} We perform stage-wise training.
Similar to previous works \citep{lu2016visual,yang2018graph} the object detector and the backbone are firstly fine tuned on VG and then frozen.
Then, the following modules are trained with different supervisions: SRF module is trained with logistic loss, and the TTST message passing module is trained with the sum of cross entropy for object classification and relation classification.
SGD ($lr=5\times10^{-3}$) is applied for optimization with momentum $0.9$. The learning rate begins to decay after the first 10 training epochs and it decays $10\%$ after each 3 epochs. The whole model is jointly trained until the loss convergences.


\textbf{Evaluation.~}
We look into three universal evaluation tasks for scene graph generation.
(1) \textbf{Predicate classification} (PredCls): given the groundtruth bounding boxes and labels of objects, predict edge (relation) labels.
(2) \textbf{Scene graph classification} (SGCls):  given groundtruth bounding boxes of objects, predict node (objects) labels and edge labels. 
(3) \textbf{Scene graph detection} (SGGen): predict boxes, node labels and edge labels given an image. SGGen is the more realistic and important metric since in practice the groudtruth bounding boxes and labels of objects are not accessible.
Only when the predicted labels of the subject, relation, and object of a relationship match the ground truth annotation, and the boxes of subject and object have more than $50\%$ IoU with the ground truth ones simultaneously, is this detection counted as correct.
The recall@K metrics ($K=[20,50,100]$) for relations  \citep{lu2016visual,yang2018graph,zellers2018neural} are used to evaluate the system performance.

\begin{table*}[t]
\caption{Ablation studies for the language prior and class-relationships prior (classification confidence) for TTST message passing as formulated in Eq.~(4) in the manuscript. We compare the effect of language prior with the one of object classification confidence.}
\label{tab:ablation_embed}
\centering
\begin{threeparttable}
\begin{tabular}{cccccccccc}
\toprule
\multirow{2}{*}{ID} & \multirow{2}{*}{ Embeddings } &  Detection &\multicolumn{2}{c}{SGGen} & \multicolumn{2}{c}{SGCls} & \multicolumn{2}{c}{ PredCls } \cr 
 \cmidrule(lr){3-3} \cmidrule(lr){4-5} \cmidrule(lr){6-7} \cmidrule(lr){8-9}
&  & mAP & R@50 & R@100 & R@50 & R@100 & R@50 & R@100\cr
\midrule

1 & - & 18.7 & 21.5 & 24.6 & 31.4 & 33.8 & 62.9 &65.1 \\
2 &Confidence   &19.7 & 25.4 & 29.6 & 33.9 & 36.2 & 65.8 &67.2 \\
3 &Language &\textbf{20.8} &\textbf{29.3} &\textbf{33.1} &\textbf{37.3} &\textbf{38.3} &\textbf{66.5} &\textbf{67.7} \\
\bottomrule
\end{tabular}
\end{threeparttable}
\end{table*}

\subsection{Quantitative Comparisons}
\label{subsec:results}
The quantitative results from different models are compared in Tab.~\ref{tab:results}.
We compare our methods with the recent strong models: \emph{MotifNet} \citep{zellers2018neural} that learns regularities using RNN, capturing context by message passing (\emph{IMP} \citep{xu2017scene}), \emph{Graph R-CNN} \citep{yang2018graph}, \emph{Mem} \citep{wang2019exploring}), Attention for Visual Relationship (\emph{AVR}) \citep{hu2019exploiting}, \emph{VRD} \citep{lu2016visual} which uses language prior, \emph{KERN}~\citep{chen2019knowledge} that exploits statistical prior knowledge and the strong frequency baseline \emph{MotifNet-Freq}~\citep{zellers2018neural}.
Because the works \emph{MSDN} \citep{li2017scene}, \emph{FacNet} \citep{li2018factorizable} and CRF-like work \emph{DRNet}~\citep{dai2017detecting} apply their own data split, we do not compare with them here.
It is worth noting that their basic object detector is reported to have $20.4\%$ in mAP@0.5 in terms of object detection accuracy ~\citep{zellers2018neural}, while our implementation of the same basic object detector has only $16.6\%$ in mAP@0.5 when we use their released code, which means that we do not have an advantage in the front-end object detector. Thus, the comparison is not in favor of ours in terms of object detection.

From Tab.~\ref{tab:results}, we observe that our final model \emph{TTST} outperforms other methods on all metrics (a little inferior to MotifNet and KERN for PredCls.
It is caused by the SRF. We will clarify the reasons in Sec.~\ref{sub:ablation}
).
It demonstrates that our method improves scene graph generation significantly.
Specifically, our method is superior to FacNet, Graph R-CNN and Mem which attempt to capture context using message passing approaches. 
Compared to VRD and NLPVR which explicitly exploits language prior, our method shows significant improvement.
It suggests that, compared to using language prior separately to predict the relationship labels, our model effectively integrates it with the visual context and learns more powerful representation.
Our method also outperforms MotifNet and the strong frequency baseline MotifNet-Freq, which indicates that our model not only learns the co-occurrence statistics of combination (subject, relation, object) from the training data but also explores the context in the given scene. 

It's worth noting that, our model \emph{TTST (w/o SRF)} without using the SRF module already achieves better results than the previous work, especially in task of PredCls. We will discuss more in Sec.~\ref{sub:ablation}. 


\subsection{Ablation Studies}
\label{sub:ablation}
Three modules are applied to boost the performance of scene graph generation: SRF, \emph{TTST} and an embedding operation of subject and object for prediction relationship (PredE).
To study how each of them affects the final performance, we perform several ablation experiments.


\subsubsection{Effectiveness of TTST} 
In Tab. \ref{tab:ablation}, Model 1 is the baseline scheme which predicts the relationship between labels by combining the features of subject, union box, and object.
Comparing Model 1 and Model 2, we find that \emph{TTST} boosts the overall performance significantly. For the SGGen setting, \emph{TTST} brings 4.4\% and 4.0\% improvement for R@50 and R@100 respectively. \emph{TTST} efficiently exploits context by the target-tailored source- transformation for message passing and enables the powerful feature representation learning.
Such visual context is clearly helpful for understanding the interaction between objects, and object detection ($1.9\%$ mAP gain). 

\subsubsection{Effectiveness of Language Context in TTST} 
We add the language prior into the \emph{TTST} in Model 3 (see Tab. \ref{tab:ablation}) as described in Eq.~\eqref{eq:stamp_obj}. This brings further significant improvement when compared with Model 2.
It demonstrates that: 1) language prior helps better explore the context among objects and relationships, and 2) \emph{TTST} effectively integrates language prior with visual context through the message passing rather than only using it as association information as in previous works, \eg, \citep{lu2016visual,li2017scene}.

\subsubsection{Effectiveness of Embedding for Inferring Relation} 
In Model 4, we additionally embed the predicted class information of subject and object for predicting their relation (PredE). We can see that the performance is further improved in all the settings. For the SGGen setting, PredE brings 4.6\% and 6.0\% in R@50 and R@100, respectively.
The explicit introduction of the semantics (class types) of the two objects involved in a relationship is helpful to infer the relations, where the co-occurrence of relationship triplet (subject, relation, object) is mined.

\subsubsection{Effectiveness of SRF} 
\label{subsec:E-SRF}
Finally, we apply SRF to prune the spurious edges to get a sparsely connected graph (Model 5). Note that when SRF is not utilized, we select 128 object pairs for the subsequent message passing based on confidence scores. Particularly, we define a confidence score for an object pair as the product of the predicted label confidences of subject and object. The object pairs with top 128 confidence scores are selected (due to the limited GPU memory).
We notice that almost all performances are improved, except that PredCls (which uses groudtruth bounding boxes and class labels) is a bit inferior to Model 4. That is because SRF may mistakenly remove some ``good"  candidates of relationship. We enable SRF to keep \emph{at most} 128 pairs. But for some images there are less pairs that meet the threshould of SRF. In contrast, the scheme without using SRF can avoid such removing.
The improvements for SGGen and SGCls demonstrates that SRF is effective in selecting the object pairs which are likely to have relationships, especially when the object proposals are noisy.
We analyze the gain taken by SRF as follows.
Even though deep learning technologies enable the network to learn powerful features from reasonable input, the learned features contain noise or interfering information, because of the imperfect model, training strategy, \etc.
If the input is preprocessed in order to remove noise or interfering information, the model is likely to learn better features.
In our model, the spurious relations between objects broadcast the interference via message passing through the graph and deteriorate the model learning process.
SRF effectively reduces such kind of interference by removing the spurious relations.


\begin{table*}[t]
\caption{Comparison of using proposed SRF to select object pairs (that are likely to have meaningful relationships) with that using motif frequency~\cite{zellers2018neural} information and the confidence score of object pair.
The experiments are conducted on our final framework.
}
\label{tab:srf_baseline}
\centering
\begin{threeparttable}
\begin{tabular}{ccccccccccc}
\toprule
\multirow{2}{*}{ ID } &\multirow{2}{*}{ Method } & \multicolumn{3}{c}{ SGGen } & \multicolumn{3}{c}{ SGCls } & \multicolumn{3}{c}{ PredCls } \cr
    \cmidrule(lr){3-5} \cmidrule(lr){6-8} \cmidrule(lr){9-11}
&& R@20 & R@50 & R@100 & R@20 & R@50 & R@100 & R@20 & R@50 & R@100\cr
\midrule
1 &  Pair Confidence & 22.4 & 29.3 & 33.1 & 34.2 & 37.3 & 38.3 & \textbf{61.8} & \textbf{66.5} & \textbf{67.7} \\
2 & Frequency & 21.0 & 28.4 & 31.1 & 32.7 & 35.1 & 35.8 & 61.5 & 66.0 & 67.6 \\
3 & SRF & \textbf{23.8} & \textbf{32.3} & \textbf{35.4} & \textbf{35.1} & \textbf{38.6} & \textbf{39.7} & 60.3 & 64.2 & 66.4 \\
\bottomrule
\end{tabular}
\end{threeparttable}
\end{table*}

\renewcommand{\arraystretch}{1.3} 
\begin{table*}[t!]
\centering
\caption{Ablation study on the influence of the number of iterations of message passing (to update the representation of nodes and edges) on the final performance. These are evaluated on our full model, which includes SRF, \emph{TTST}, Language and PredE.}
\label{tab:ablation_iteration}
\begin{threeparttable}
\begin{tabular}{ccccccccccc}
\toprule
\multirow{2}{*}{ IteNr. } & Object Detection  & \multicolumn{3}{c}{ SGGen } & \multicolumn{3}{c}{ SGCls } & \multicolumn{3}{c}{ PredCls } \cr
    \cmidrule(lr){3-5} \cmidrule(lr){6-8} \cmidrule(lr){9-11}
& mAP & R@20 & R@50 & R@100 & R@20 & R@50 & R@100 & R@20 & R@50 & R@100\cr
\midrule
1 & 19.1 & 19.3 &26.5 & 30.1 & 29.8 & 32.3 & 34.7 & 57.6 &59.8 & 62.1\\
2 & \textbf{20.8} &\textbf{23.8} & \textbf{32.3} & \textbf{35.4} & \textbf{35.1} & \textbf{38.6} & 39.7 & 60.3 & 64.2 & 66.4 \\
3  & 20.6 &23.1 & 32.2 & 35.4 & 34.4 & 37.1 & \textbf{40.1} & \textbf{60.5} & \textbf{65.7} & \textbf{67.8} \\
\bottomrule
\end{tabular}
\end{threeparttable}
\end{table*}
\subsubsection{Language Prior v.s. Class-relationships Prior}

It is a popular method that explores the class-relationships prior by utilizing object classification confidence for scene graph generation \citep{dai2017detecting,yang2018graph}.
In order to compare the effectiveness of language priors and class-relationships prior in our \emph{TTST}, we design this ablation studies by replacing the embedding $e_i$ in Eq. \eqref{eq:stamp_obj} by the corresponding classification confidence $p_i^o$. 
The results are given in Tab.~\ref{tab:ablation_embed}.

We can see that, both language prior and class-relationships prior help our \emph{TTST} module to get better performance. The class information plays an important role in guiding the message passing in \emph{TTST}.
However, using language prior to explore the context brings more improvements than the one using class-relationships prior.
Class-relationships prior provides the contextual information of the co-occurrence of different object classes within an image in the training dataset.
Language prior contained in the learned word2vector does not only reflect the  co-occurrence of the nouns within a sentence but also the semantic relations in the language space. For example, ``man" and ``boy" have different class-relationships with ``bike", but their distances to ``bike" in the word2vector space are similar, because ``man" and ``boy" share the similar semantic meaning ``human being".

\subsubsection{Effectiveness of SRF v.s. Frequency-based Method}

Previous works have proposed different methods to remove redundant object pairs to reduce computation and improve performance w.r.t. visual relationship detection~\citep{li2017vip,dai2017detecting,li2017scene,yang2018graph}.
In this paper, we propose SRF to filter out the object pairs that are unlikely to have relationships, by utilizing language prior.
Zellers \etal \citep{zellers2018neural} propose a strong frequency baseline for visual relationships.
This statistical information between different object classes and relation types is worthy being explored to select object pairs that are likely to have meaningful relationships.
Therefore, we conduct an ablation study to compare the effectiveness of SRF, which is trained to select object pairs using language prior, with the one that uses frequency information.
When SRF is removed, we select the 128 object pairs which have the top pair confidence scores that equals the multiplication of the object classification scores of the \emph{subject} and \emph{subject}.
Tab.~\ref{tab:srf_baseline} shows the results.
In the motif frequency setting, 128 object pairs are selected (which are the most frequently occurring in the training set and their corresponding frequency is larger than $0.01$).

We can see that SRF outperforms the other two settings in the tasks of \emph{SGGen} and \emph{SGCls}.
SRF is a little inferior to motif frequency for PredCls with the similar reason as explained in the previous section \ref{subsec:E-SRF}. 
Under the PredCls setting, all objects are groundtruth and there are very few object pairs that have no relationship. Selection based on pair confidence and motif frequency (128 pairs) retain almost all object pairs but SRF would mistakenly remove some (less than 128 pairs).
Pair confidence settings perfrom the best in PredCls because it keeps almost all object pairs (because the confidence score of each object is 1) while motif frequency removes the object pairs with low statistic frequency.


In SGGen, there are a large number of object pairs (because of many predicted object candidates), selection with frequency will only retain the very frequent relationships and the not so frequent ones are removed.
In contrast, SRF is able to effectively determine which object pairs have semantic relationships even though they are rare in the training set.
On the other hand, selection with frequency needs to adapt the object classification results of faster r-cnn whose accuracy is insufficient. This leads to false removal of the object pairs.
In contrast, SRF uses the semantics learned via embedding to reduce the negative effects of inaccurate object classification.
This is also the reason why SRF outperforms the selection with frequency and pair confidence in SGGen and SGCls.

\renewcommand{\arraystretch}{1.3} 
\begin{table*}[t]
\centering
\caption{Ablation study of how different message passing directions in the \emph{TTST} modules affect the performance.
``rel-obj" denotes passing message from relationship to object, and the other notation are similar. The full model is implemented.}
\label{tab:ablation_obje_region}
\begin{threeparttable}
\begin{tabular}{c|cccc|cccccccc}
\toprule
\multirow{2}{*}{Model} &  & &  & & Detection &\multicolumn{2}{c}{SGGen} & \multicolumn{2}{c}{SGCls} & \multicolumn{2}{c}{ PredCls } \cr 
  \cmidrule(lr){6-6} \cmidrule(lr){7-8} \cmidrule(lr){9-10} \cmidrule(lr){11-12}
&obj-obj  &rel-obj  &obj-rel  &rel-rel  & mAP & R@50 & R@100 & R@50 & R@100 & R@50 & R@100\cr
\midrule
0 & - & - &- &-						& 16.6 & 14.1 &18.5 &27.7 & 30.5 &54.2 &58.4 \\
1 & - & - &\checkmark & -   	    & 16.7 & 14.8 &19.7 &28.7 & 32.0 &59.7 &62.8 \\
2 & - & - &- &\checkmark 	        & 16.6 & 14.7 &19.6 &27.9 & 31.4 &58.2 &61.3 \\
3 & - & - &\checkmark &\checkmark 	& 16.7 & 15.1 &20.2 &29.0 & 32.3 &60.9 &63.4 \\
4 &\checkmark & - &- &- 			& 19.8 & 25.5 &28.1 &32.5 & 36.8 &55.8 &59.7 \\
5 &- &\checkmark & - 		& - 	& 16.8 & 14.2 &18.5 &28.0 & 31.1 &55.5 &59.5 \\
6 &\checkmark &\checkmark &- & -   	& 20.2 & 26.4 &28.6 &33.8 & 37.7 &56.1 &59.8 \\
7 &\checkmark &\checkmark &\checkmark &\checkmark &\textbf{20.8} & \textbf{32.3} & \textbf{35.4} & \textbf{38.6} & \textbf{39.7}  & \textbf{64.2} & \textbf{66.4} \\
\bottomrule
\end{tabular}
\end{threeparttable}
\end{table*}

\subsubsection{Iteration of Message Passing} 
\emph{TTST} works in an iterative way to update the representation of nodes and edges, it is necessary to study how different numbers of iterations affect the final performance.
Our full model is trained in different iterations of message passing and reports the results in Tab. \ref{tab:ablation_iteration}.
We notice that the overall performance increases with more iterations of message passing and most of the performance reaches the best after 2 iterations.
After 3 iterations, some performances drop, especially object detection. But the performance in the PredCls task setting is still slightly improved. We analyze the reason as follows.

The context is captured by passing message to the neighbors via \emph{TTST}.
In one iteration the message is broadcast to its neighboring nodes.
More iterations will broadcast the message to further nodes (edges) and capture wider context. 
Therefore, overall performance is improved. 
However, the noise and interfering information are also broadcast through the graph.
With more iterations, each node/edge accumulates such harmful information in parallel with collecting context from others.
Normally, a graph for an image is not large and information will go through the graph within 2 steps starting from any node (see Fig. \ref{fig:qualitative}).
Thus, the context is already extracted sufficiently in two iterations and harmful information keeps accumulating with more iterations.
Consequently, the model performance becomes worse with the deteriorated performance in object detection.
Other existing works which pass messages iteratively also reported similar problem \citep{li2017scene,liu2018structure,li2018factorizable}. 
However, slight gains are obtained for PredCls which isolates the performance of object detection.
It is because the relation representations are more complex and difficult than that of an individual object.
More iterations will help refine the relation representations.
The weaker object detection decreases the performance of SGGen and SGCls. 
Based on this study, we use 2 iterations in our final scheme.

\subsubsection{Message Passing Direction}
As formulated in Eq.~\eqref{eq:stamp_obj}\eqref{eq:stamp_rel}, message are propagated in four directions in \emph{TTST} modules: obj-obj,  rel-obj, obj-rel and rel-rel.
We evaluated how each of the message passing direction affects the performance of our model.
The results presented in Tab. \ref{tab:ablation_obje_region} shows that any direction of message passing improves the performance of the framework (compared with Model 0) and the full message passing model has the best performance (Model 7).
By comparing Model 1-3 with Model 4-6 correspondingly, we notice that passing message to objects (Eq.~\eqref{eq:stamp_obj}) improves the performance of object detection (mAP is improved from $16.6\%$ to $20.2\%$).
Consequently, the performance of SGGen (from $14.1\%$ to $26.4\%$) and SGCls (from $27.2\%$ to $33.8\%$) are improved significantly.
When information is propagated to relationships (see Eq.~\eqref{eq:stamp_rel}), the performance of PredCls is improved from $54.2\%$ to $60.9\%$ ($12.4\%$ relative gain).
Consequently, the overall performance is improved.
Compared with those improvements from *-obj and *-rel separately, the full model shows further overall improvement.
It demonstrates that the \emph{TTST} modules learn the context by propagating information among objects and relationships effectively and benefit the mutual promotion of object detection and relationship detection.

\subsubsection{Improvements on Object Detection}
As shown in Tab.~\ref{tab:ablation} and Tab.~\ref{tab:ablation_obje_region}, \emph{TTST} modules not only improve the performance of visual relationship detection but also the performance of object detection, which is one of the most important tasks for visual scene understanding and critically affects the overall performance of scene graph generation. We achieve the goal of mutual promotion of visual relationship detection and object detection.

\begin{figure*}[t!]
    \centering
    \includegraphics[width=1.0\linewidth]{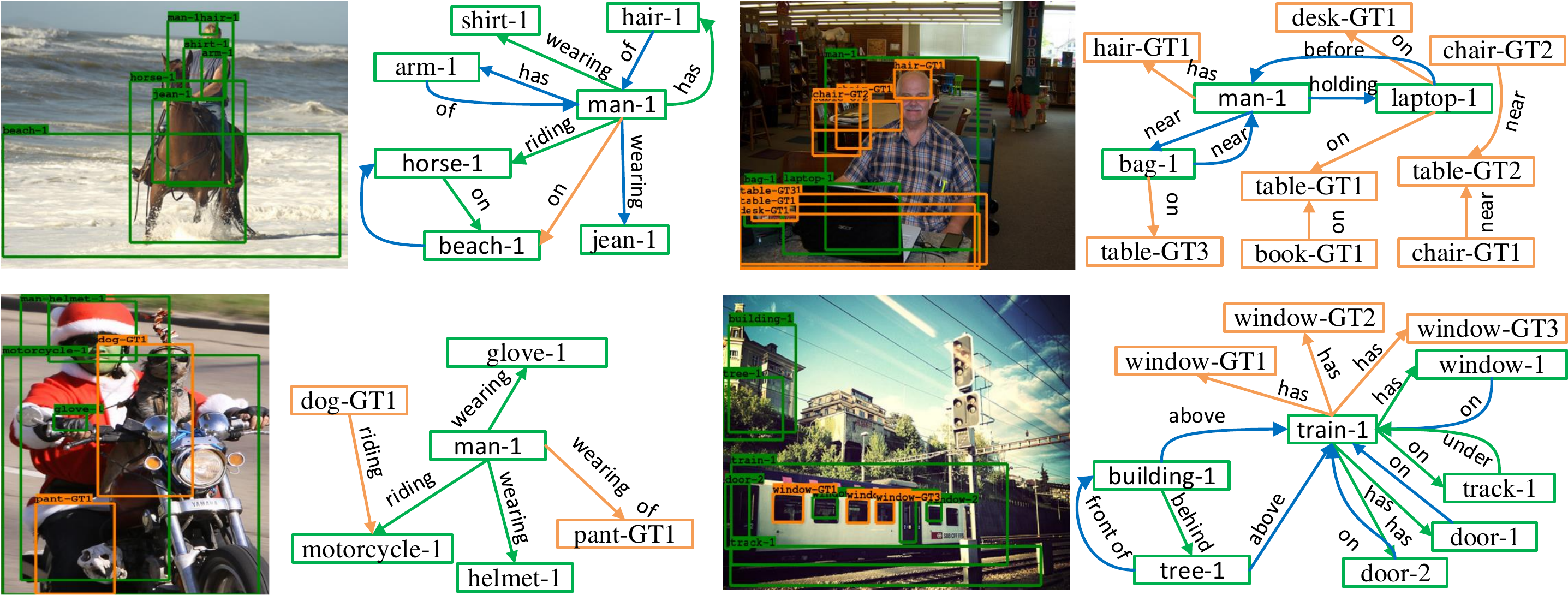}
    \includegraphics[width=1.0\linewidth]{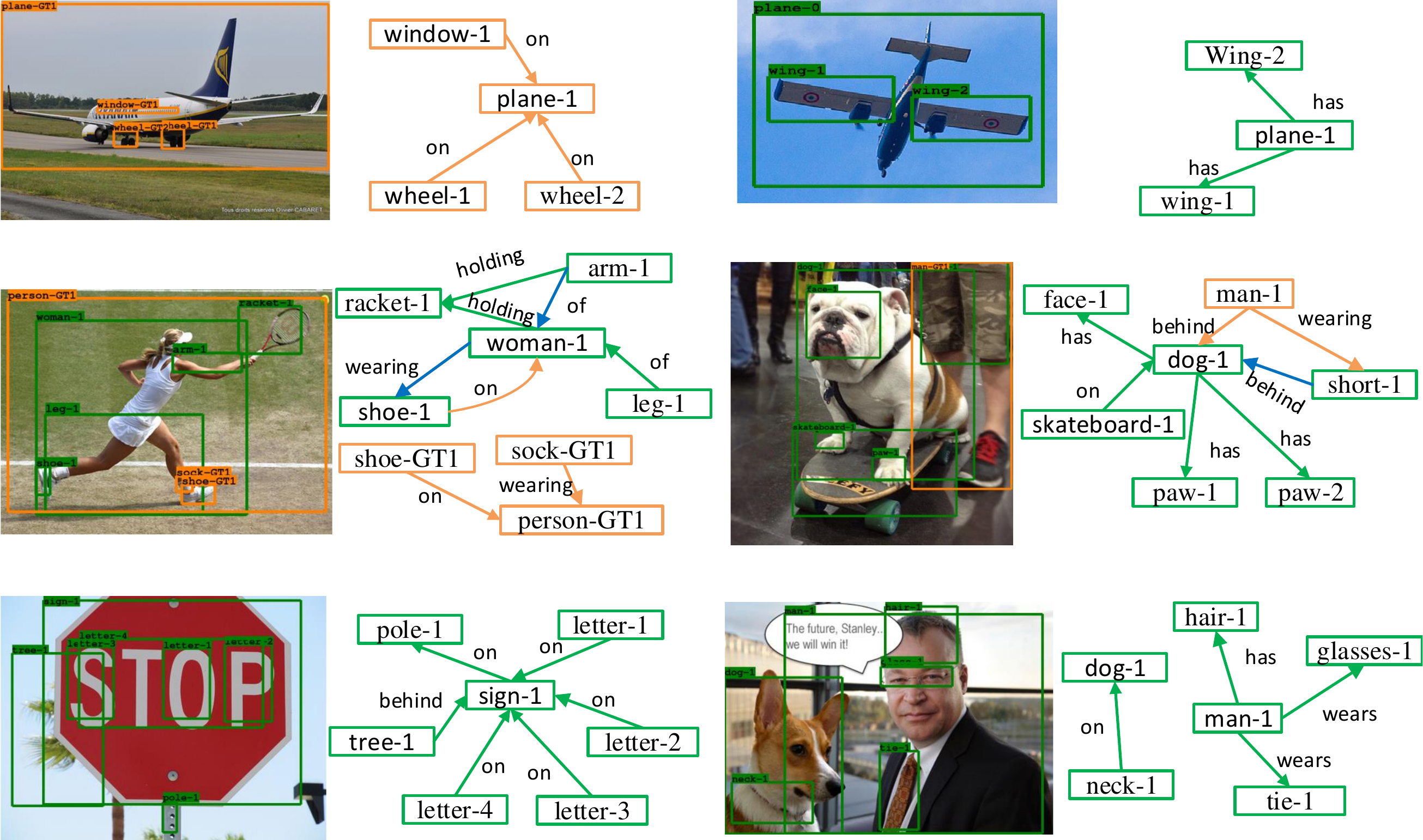}
    \caption{Qualitative results from our model in the scene graph generation setting. 
    Green boxes denote the correctly detected objects while orange boxes denote the ground truth objects that are not detected.
    Green edges correspond to the correctly recognized relationships at the R@20 setting while orange edges denote the ground truth relationships that are not recognized. The blue edges denote the recognized relationships that however do not exist in the ground truth annotations.
    }
    \label{fig:qualitative}
\end{figure*}

\subsection{Qualitative Results}
Fig. \ref{fig:qualitative} shows scene graphs generated by our model from the test set.
We can see that our model is able to infer relationships between object pairs correctly (green edges) and generate high-quality scene graphs.
Some true relationships that are not annotated in the ground truth also can be inferred correctly (blue edges), \textit{e.g.}``man-wearing-jeans" in the first image.
It implies that our model works even better than what the quantitative results demonstrate because the unannotated but correctly predicted relationships would deteriorate the performance under current evaluation metrics.

From the examples, we notice that when the detector fails, all the inference of edges to the object will be false, and this situation often occurs when detecting small objects.
For example in the right image of the first row, many small or occluded objects are not correctly detected (orange boxes) and all edges connecting them are not recognized correctly.
Another common failure case is caused by the ambiguity of relation types, \textit{e.g.} ''wear" vs. ``wearing".

\section{Conclusion}
\label{sec:conclusions}
This paper proposes a novel and effective target-tailored source-transformation (\emph{TTST}) for message passing to generate scene graph.
Our model includes a SRF that effectively prunes the spurious connections between objects, and \emph{TTST} modules that learn context by simultaneously ``seeing" the target and source objects.
Language prior is used to help message passing and integrated with visual context to learn powerful representations.
The experimental results show that our method significantly outperforms the state-of-the-art methods for scene graph generation and meanwhile the performance of object detection is improved.
The extensive ablation studies demonstrate the contribution of each proposed module to the framework.

\bibliographystyle{spbasic}
\bibliography{ref}

\end{document}